\documentclass[11pt]{article}
\usepackage{eacl2017}
\usepackage{times}
\usepackage{url}
\usepackage{latexsym}
\usepackage{graphicx}

\RequirePackage{graphicx}
\RequirePackage{amsmath,amsfonts,amssymb}
\allowdisplaybreaks

\RequirePackage{etoolbox,balance}
\RequirePackage{booktabs,makecell,multirow,array,colortbl,dcolumn,stfloats}
\RequirePackage{xspace,xstring}
\RequirePackage{footmisc}
\RequirePackage[svgnames,dvipsnames]{xcolor}
\RequirePackage{iftex}
\RequirePackage{microtype}
\usepackage{cuted}

\eaclfinalcopy %

\title{A simple language-agnostic yet very strong baseline system for hate speech and offensive content identification}

\author{Yves Bestgen \\
  Laboratoire d'analyse statistique des textes - Statistical Analysis of Text Laboratory (LAST - SATLab) \\
  Institut de recherche en sciences psychologiques \\
  Universit\'e catholique de Louvain \\
  Place Cardinal Mercier, 10 1348 Louvain-la-Neuve, Belgium \\
  \texttt{yves.bestgen@uclouvain.be} \\}

\date{}

\begin{document}
\maketitle
\begin{abstract}
For automatically identifying hate speech and offensive content in tweets, a system based on a classical supervised algorithm only fed with character n-grams, and thus completely language-agnostic, is proposed by the SATLab team. After its optimization in terms of the feature weighting and the classifier parameters, it reached, in the multilingual HASOC 2021 challenge, a medium performance level in English, the language for which it is easy to develop deep learning approaches relying on many external linguistic resources, but a far better level for the two less resourced language, Hindi and Marathi. It ends even first when performances are averaged over the three tasks in these languages, outperforming many deep learning approaches. These performances suggest that it is an interesting reference level to evaluate the benefits of using more complex approaches such as deep learning or taking into account complementary resources. 
\end{abstract}
The diffusion of hate speech and offensive content in social networks has become a crucial problem. The tremendous number of posts broadcasted at any given time prevents their identification by human evaluation. This task is made even more complex by the large number of languages in which these offensive contents are spread. Not surprisingly, a lot of research is being done to develop automatic detection systems. As in many NLP domains, deep learning approaches and the use of pre-computed embeddings have proven to be the most efficient, even in languages with few resources \cite{DBLP:conf/fire/0001MMPDMP19,DBLP:conf/fire/0001MMC20}. However, traditional machine learning systems have sometimes proven to be very competitive \cite{DBLP:conf/fire/MujadiaMS19,DBLP:conf/fire/SarojMP19}. One may thus wonder what level of performance can be achieved by a much simpler yet heavily optimized classical supervised approach, completely language-agnostic, based only on a few thousand examples to feed the supervised learner but without any additional resources. If this system is (relatively) successful, it would give a computationally easy baseline that could help evaluating the benefits of additional knowledge, complex architectures, deep learning or language expertise.

The HASOC 2021 shared task "Hate Speech and Offensive Content Identification in English and Indo-Aryan Languages" \cite{hasoc2021mergeoverview} is particularly relevant for developing such a system because it proposes three languages. Among them, one, English, is obviously the most studied language in automatic language processing and the one in which the largest number of resources is available. Hindi and, even more so, Marathi have been much less studied and are still classified as low-resource languages \cite{ws-2018-deep,OrtegaProceedings,gaikwad2021cross}. One can think a priori that the approach proposed here will be much more competitive in these two languages. 

The remainder of this paper presents the datasets made available for this shared task and the challenge rules, the system developed, and the results obtained which confirms that the proposed approach is a strong language-agnostic baseline for hate speech and offensive content identification.

\section{Materials and Task}

The SATLab participated in subtask 1 of the HASOC 2021 shared task "Hate Speech and Offensive Content Identification in English and Indo-Aryan Languages" which proposes two problems to be solved in three languages \cite{hasoc2021mergeoverview}. The first problem requires to categorize tweets into two categories: Hate and Offensive (HOF) or not (NOT). It is proposed for English, Hindi and Marathi. The second problem requires categorizing the same tweets into four categories, by dividing the Hate and Offensive category into three subcategories: Hate speech (HATE), Offensive (OFFN) and Profane (PRFN). It is offered for English and Hindi.

For each language, learning and test materials have been provided by the task organizers \cite{hasoc2021overview,gaikwad2021cross}. The frequencies (\#) and percentages (\%) in each category of each problem for each language are given in Table 1.

\begin{table*}
\centering
\caption{Dataset statistics of subtask 1}
\label{tab:data}
\begin{tabular}{lcrrrrrrrrrr}
\toprule
   &   & \multicolumn{8}{c}{Learning Phase} &\multicolumn{1}{c}{Learning}& \multicolumn{1}{c}{Test} \\ 
   &   & \multicolumn{2}{c}{Problem 1} && \multicolumn{4}{c}{Problem 2} & & \multicolumn{1}{c}{Phase} & \multicolumn{1}{c}{Phase} \\ \cmidrule{3-4} \cmidrule{6-9} \cmidrule{11-12}
&  & \multicolumn{1}{c}{NOT} & \multicolumn{1}{c}{HOF} && \multicolumn{1}{c}{NONE} & \multicolumn{1}{c}{HATE} & \multicolumn{1}{c}{OFFN} & \multicolumn{1}{c}{PRFN} && \multicolumn{1}{c}{Total} & \multicolumn{1}{c}{Total} \\ 
\midrule
English & \# & 1342 & 2501 && 1342 & 683 & 622 & 1196 && 3843 & 1281 \\ 
 & \% & 34.9 & 65.1 && 34.9 & 17.8 & 16.2 & 31.1 && 75.0 & 25.0 \\  
Hindi & \# & 3161 & 1433 && 3161 & 566 & 654 & 213 && 4594 & 1532  \\  
 & \% & 68.8 & 31.2 && 68.8 & 12.3 & 14.2 & 4.6 && 75.0 & 25.0  \\  
Marathi & \# & 1205 & 669 &&       &     &     &     && 1874 & 525  \\  
 & \%  & 64.3 & 35.7 &&     &      &     &     && 78.1 & 21.9  \\  
\bottomrule
\end{tabular}
\end{table*}

This table deserves several comments. First of all, the learning set is much smaller in Marathi (18\% of the total) than in the other two languages, the difference between the two latter being much smaller (37\% of the total in English and 45\% of the total in Hindi). The proportion of tweets in the HOF category is much larger in English than in the other two languages. The difference clearly comes from the PRFN category which is much more frequent in English than in Hindi where it represents only a very small percentage. 

\subsection{Challenge rules}
The rules of the challenge allowed teams to use any additional resources including materials from previous HASOC tasks, lexical norms such as emotional word lists, precomputed embeddings, the use of syntactic parsers or even machine translation systems to analyze other languages in English. The system proposed by the SATLab does not include any of these additional resources.

The official measure chosen by the organizers to rank the teams in the challenge is the Macro-F1 which has the advantage of giving the same weight to all categories, however rare they may be (e.g., less than 5\% of PRFN in Hindi). 

Each team was allowed to submit five runs for each subtask between August 20 and 30, 2021, and the team's best performance was displayed in the Leaderboard. Compared to the ten or so other shared tasks I participated in, it is important to underline that the submission system proposed by the challenge organizers (\url{https://hasocfire.github.io/submission/login.html}) was particularly ergonomic. Moreover, the fact that the teams could not hide their best score, as it is often the case in other systems, made, in my opinion, the competition more fair. 

\section{Proposed System}

In order to meet the requirements presented in the introduction, the proposed system is only based on character n-grams \cite{bestgen:2017:VarDial}, an approach frequently used in automatic language processing when the developed system has to support several languages. These n-grams were extracted from the lowercased tweets with the only specificity that those starting or ending the tweet were distinguished from the others by the presence of a specific character. All character n-grams observed at least twice in the material were retained.

During the n-gram extraction, three parameters had to be set:  
\begin{itemize}
\item The length of the n-grams in number of characters. The minimum length was systematically set to 1 while the maximum lengths evaluated varied between four and eight characters.
\item The weighting applied to the frequency of each feature in each instance. Two well-established weighting schema were evaluated:
\begin{itemize}
 \item Sublinear TF-IDF:
\begin{strip}
\begin{equation}
   \text{(sl)TF-IDF} = (1 + \log(tf)) \times \log\frac{N}{df}
\end{equation}
\end{strip}
where $tf$ refers to the frequency of the term in the document, $N$ is the number of documents in the set and $df$ the number of documents that include the term.
 \item BM25 (\cite{Robertson:2009:PRF,bestgen:2021:VarDial}), which is considered as one of the most efficient weighting schema \cite{book:manning:2008}. It is a kind of TF-IDF that takes into account the length of the document. The following formula was used: 
\begin{strip}
\begin{equation}
   \text{BM25} = \frac{tf}{tf + k_1 * (1 - b + b * \frac{dl}{dl-avg_{dl}})} \times\log\frac{N - df + 0.5}{df + 0.5} 
\end{equation}
\end{strip}
in which
\begin{itemize}
\item $\frac{tf}{tf + k_1}$ is the TF component which, contrarily to the usual TF-IDF, has an asymptotic maximum tuned by the $k_1$ parameter.
\item $(1 - b + b * \frac{dl}{dl-avg_{dl}})$, where $dl$ is the length of the document and $avg_{dl}$, the average length of the documents in the set, is the document length normalization factor whose impact is tuned by parameter $b$ (and by $k_1$).
\item The second part of the formula is a variant of the usual IDF, proposed by Robertson and Sp\"{a}rck Jones \cite{Robertson:2009:PRF}.
\end{itemize}
In our analyses, $k_1$ was set to 2 and $b$ to 0.75.
\end{itemize}
\item Normalization of the feature scores for each instance:
\begin{itemize}
 \item The classical L2 regularization.
 \item A MinMax transformation: 
 \begin{strip}
 \begin{equation}
    \text{MinMax} = \frac{Feature_i\_score - min} {max - min} + 0.01
\end{equation} 
\end{strip}
It is important to note that this transformation is applied independently to each instance and not, as is often the case, to each feature. The value of 0.01 is added to distinguish the lowest scoring feature of an instance with the value of 0, which codes the absence of a feature. 
\end{itemize}
\end{itemize}

These character n-grams were the only features provided to the supervised learning procedure. Two well-established procedures were evaluated: 
\begin{itemize}
\item The (dual) L2-regularized logistic regression as implemented in the LIBLinear package \cite{Fan2008}, an extremely fast approach and very simple to use because it only requires the optimization of two parameters. The two parameters to optimize are the regularization parameter C and -wi which allows to adjust this parameter C for the different categories. This approach was used for the initial submission to each of the five problems.
\item A much slower and more complex approach to optimize because it requires the optimization of many parameters, but that has recently outperformed all deep-learning based systems participating in the CMCL 2021 shared task on predicting gaze data during reading \cite{bestgen-2021-last}: a gradient boosting decision tree approach as implemented in the LightGBM free software \cite{LightGBM}. This approach has been used only in a second time.
\end{itemize}

The system was independently optimized for each language during the learning phase using a 3-fold cross-validation procedure, whose folds were stratified according to the four categories of problem 2 for English and Hindi and the two categories of problem 1 for Marathi. This cross-validation step led to setting the parameters described above as shown in Table 2 for the initial SATLab submissions.

\begin{table*}
\centering
\caption{Parameters for the initial submissions}
\label{tab:param}
\begin{tabular}{lccccc}
\toprule
Language & \multicolumn{2}{c}{English} & \multicolumn{2}{c}{Hindi} & \multicolumn{1}{c}{Marathi} \\ 
Problem    &     1   &   2   &  1   &  2   &   1 \\ 
\midrule
 N-gram length  & 5   &   5  &   5   &  5    &  5 \\ 
 Weighting   &  TF-IDF & TF-IDF & BM25 & TF-IDF & BM25 \\ 
 Normalization & MinMax  & L2  &  L2 & MinMax & L2 \\ 
 C           &    1.1  & 2.5  & 3.7  & 0.083  &  6 \\ 
 w\_HOF        &    0.5  &      & 2.2  &       &  2 \\ 
 w\_HATE          &         & 2.0  &      & 1.87 & \\ 
 w\_OFFN          &         & 3.0  &      &  0.93 & \\ 
 w\_PRFN         &         & 0.8  &      & 5.60  &    \\ 
\bottomrule
\end{tabular}
\end{table*}

\section{Results}
In this section, the performance of the initial system proposed by the SATLab and the various optimization attempts that have been made are first presented. Secondly, these performances are compared to those of other teams in order to determine whether the proposed approach is competitive enough to serve as a baseline for evaluating the benefits of using deep learning approaches and resources supplementary to those provided in the task itself.

\begin{table*}
\centering
\caption{Macro-F1 during cross-validation and on the test set}
\label{tab:res1}
\begin{tabular}{lccccc}
\toprule
Language & \multicolumn{2}{c}{English} & \multicolumn{2}{c}{Hindi} & \multicolumn{1}{c}{Marathi} \\ 
Problem    &     1   &   2   &  1   &  2   &   1 \\ 
\midrule
CV & 0.7483 & 0.5876 & 0.7551 & 0.5133 & 0.8565 \\
\midrule
Initial  & 0.7635 & \bf 0.6114 & \bf 0.7718 & 0.5563 & 0.8547 \\
Best LR  &        &        &        & \bf 0.5586 & 0.8595 \\
Best LGBM & \bf 0.7823  &      &        &        & \bf 0.8749 \\

\bottomrule
\end{tabular}
\end{table*}

\subsection{SATLab submissions} 
Table 3 presents the performance of the main versions of the SATLab system submitted for the five problems and thus the benefits brought by the optimization attempts on the test set. The first row reports the performance of the original system for each problem during the cross-validation step. Logically, the performances are less good for problems requiring the identification of more than two categories as well as when a category is particularly rare (Hindi-2). We also observe strong differences between the three languages. Since only one split into three folds was used, one can assume that these scores are, at least slightly, overestimated. 

The second row shows the performance of the same versions on the test set and thus the initial submissions to the challenge. All scores are higher on the test set than during the cross-validation step.

As it was allowed to submit five runs for each problem, I first tried to optimize the classifier based on logistic regression by modifying very slightly the two LIBLinear parameters (i.e., C and -wi). These attempts brought a (very) slight benefit for two of the five problems as shown in the third row of Table 3. 

In a second step, an LightGBM classifier was trained using a random grid search procedure for each of the five problems to try optimizing the parameters. As shown in the fourth row of Table 3, this step resulted in a stronger performance improvement in two problems: English-1 and Marathi. For the other three problems, LightGBM did not improve the performance of LIBLinear. The selected parameters for the two successful problems are given in Appendix 1. The number of boosting iterations was determined during cross-validation by using the LightGBM early stopping procedure which stops training when the performance on the validation fold doesn't improve in the last 200 rounds. The final system values on the test set for the five problems are bolded in the table. The run names of these solutions in the official leaderboard are respectively: English 1b, English 2, Hindi 1, Hindi B S4 and Marathi 3.

\begin{table*}
\centering
\caption{Macro-F1 on the test set for the five problems}
\label{tab:res1}
\begin{tabular}{rlcrlc}
\toprule
\multicolumn{3}{c}{English-1 N=56} & \multicolumn{3}{c}{English-2 N=37} \\ 
Rank    &  Team  &   Macro-F1  &  Rank    &  Team  &   Macro-F1 \\ 
\midrule
1 & NLP-CIC	& 0.8305 & 1 & NLP-CIC	& 0.6657 \\
2 & HUNLP	& 0.8215 & 2 & neuro-utmn-thales	& 0.6577 \\
3 & neuro-utmn-thales	& 0.8199 & 3 & HASOC21rub	& 0.6482 \\
... & & & ... & & \\
22 & hate-busters	& 0.7894 & 15 & KuiYongyi	& 0.6116 \\
\bf 23 & \bf SATLab	&	0.7823 & \bf 16 & \bf SATLab	& 0.6114 \\
24 & TAD	& 0.7776 & 17 & hate-busters	& 0.6096 \\
... & & & ... & & \\
\bottomrule
\multicolumn{3}{c}{Hindi-1 N=34} & \multicolumn{3}{c}{Marathi N=25} \\ 
Rank    &  Team  &   Macro-F1  &  Rank    &  Team  &   Macro-F1 \\ 
\midrule
1 & t1	& 0.7825 & 1 & WLV-RIT	& 0.9144 \\
2 & Super Mario	& 0.7797 & 2 & neuro-utmn-thales	& 0.8808 \\
3 & Hasnuhana	& 0.7797 & 3 & Hasnuhana	& 0.8756 \\
... & & & \bf 4 & \bf SATLab	& 0.8749 \\
6 & KuiYongyi	& 0.7725 & 5 &  PreCog IIIT &	0.8734 \\
\bf 7 & \bf SATLab	& 0.7718 & ... & & \\
8 & neuro-utmn-thales & 0.7682  \\
... & & & & & \\
\bottomrule
\multicolumn{3}{c}{Hindi-2 N=24} & & & \\ 
Rank    &  Team  &   Macro-F1  &      &    &   \\ 
\midrule
1 & NeuralSpace	& 0.5603 & & & \\
\bf 2 & \bf SATLab	&	0.5586 & & & \\
3 & hate-busters	& 0.5582 & & & \\
... & & \\
\bottomrule
\end{tabular}
\end{table*} 

\begin{table*}
\centering
  \caption{Transformed Macro-F1 for the five problems}
  \label{tab:meanall}
  \begin{tabular}{rlrrm{1cm}rrrr}
    \toprule
 &  & \multicolumn{5}{c}{Nbr. of problems the team participated in} \\ 
Rank & Team & \multicolumn{1}{c}{5} & \multicolumn{1}{c}{4} & \multicolumn{1}{c}{3} & \multicolumn{1}{c}{2} & \multicolumn{1}{c}{1} \\ 
\midrule
1 & WLV-RIT  & & & & & 1.0000 \\ 
2 & NLP-CIC  & 0.9814  & & & & \\ 
3 & neuro-utmn-thales  & & 0.9800  & & & \\ 
4 & HASOC21rub  & & & & 0.9693  &  \\ 
5 & NeuralSpace  & 0.9666  & & & &  \\ 
\bf 6 & \bf SATLab  & 0.9601  & & & &  \\ 
7 & KuiYongyi  & 0.9596  & & & &  \\ 
8 & CAROLL Passau  & & & & & 0.9590 \\
9 & IMS-SINAI  & & & & & 0.9569 \\
10 & hate-busters  & 0.9517  & & & & \\
... & & & & & & \\
\multicolumn{2}{c}{Number of Teams} & \multicolumn{1}{c}{16} & \multicolumn{1}{c}{8} & \multicolumn{1}{c}{6} & \multicolumn{1}{c}{18} & \multicolumn{1}{c}{15} \\
\bottomrule
\end{tabular}
\end{table*}

\begin{table*}
  \caption{Transformed Macro-F1 for the two English problems and for the Hindi and Marathi problems}
  \label{tab:meanen}
  \begin{tabular}{rlrrrlrrr}
    \toprule
\multicolumn{4}{c}{English 1 \& 2} & \multicolumn{5}{c}{Hindi 1 \& 2 and Marathi}\\ 
 &  & \multicolumn{2}{c}{\#problems}  & &  & \multicolumn{3}{c}{\#problems}\\ 
Rk & Team & \multicolumn{1}{c}{2} & \multicolumn{1}{c}{1} & Rk & Team & \multicolumn{1}{c}{3}  & \multicolumn{1}{c}{2} & \multicolumn{1}{c}{1}\\ 
\midrule
1 & NLP-CIC  & 1.0000  &  & 1 & WLV-RIT  & & & 1.0000 \\ 
2 & neuro-utmn-thales  & 0.9876  & & \bf 2 & \bf SATLab  & 0.9800  & & \\ 
3 & HASOC21rub  & 0.9693  &  & 3 & NeuralSpace  & 0.9787  & &  \\ 
4 & HUNLP  & 0.9675  &  & 4 & neuro-utmn-thales  & & 0.9725  &  \\  
5 & HNLP  & 0.9674  &  & 5 & KuiYongyi  & 0.9707  & &  \\
... & & & & 6 & NLP-CIC  & 0.9690  & & \\
19 & hate-busters  & 0.9331  &  & 7 & hate-busters  & 0.9640  & & \\ 
\bf 20 & \bf SATLab  & 0.9302  &  & 8 &  CAROLL Passau  & & & 0.9590   \\
21 & TeamOulu  & & 0.9272 & 9 & BIU  & & 0.9484  & \\ 
... & & & & ... & & & & \\
\multicolumn{2}{c}{Number of Teams} & \multicolumn{1}{c}{38} & \multicolumn{1}{c}{25} & \multicolumn{2}{c}{Number of Teams} & \multicolumn{1}{c}{17} & \multicolumn{1}{c}{13} & \multicolumn{1}{c}{8} \\
\bottomrule
\end{tabular}
\end{table*}

\subsection{Benchmarking the approach}
The main objective of SATLab's participation in HASOC 2021 was to propose a competitive system relying only on the training data and employing only classical supervised learning procedures. To determine whether this goal was achieved, Tables 4-6 compare the performance of the approach to that of the other participating systems. 

Table 4 shows for each of the five problems the number of teams that participated, the scores of the top three teams, the scores of the best SATLab version, and the scores of the two contiguous teams. As it can be seen, it is clearly in the two less resourced languages, Hindi and Marathi, that the performance of the approach is among the best since it is even second, very close to the first (and the third), in the Hindi-2 problem. In English on the other hand, the system is ranked in the middle of the pack of average scores at 0.048 and 0.054 of the best team.

The difference in performance between English and the other two languages is particularly evident in Tables 5 and 6, which present the average scores of the teams for the five problems (Table 5), the two problems in English and the three problems in less endowed languages (Table 6). Before calculating these averages, the scores for each problem were divided by the maximum score obtained for the problem in question. This transformation\footnote{This transformation of the scores considers that the minimum score in each task is the same, 0, and that therefore no correction should be made at this level. This seems to me justified by the fact that, even if it is unlikely, a system can be wrong on all instances, but also and especially because it is the deviation from the maximum score that is important.} allows to give an equivalent weight to the scores of all problems. It is then possible to present in the same table, without distorting the results, all the teams, whatever the number of problems they have participated in. Without this transformation, the teams that participated in the most difficult tasks are penalized compared to those that did not. In these tables, the number of problems each team participated in is given by the column in which the score is found and the total number of teams that participated in a given number of problems is presented in the last row. 

In terms of the overall average (Table 5), SATLab ranks sixth overall and third among the 16 teams that participated in the five tasks. In English (Table 6), on the other hand, it ranks only 20th. In the two less endowed languages (Table 6), it is second, exceeded only by a team that participated in only one of the five tasks.

\section{Conclusion}
A system, based exclusively on the character n-grams present in the posts to be categorized, employing no additional linguistic resources and thus completely language-agnostic, is proposed to automatically identify hate speech and offensive content in social network posts. It relies on traditional machine learning procedures such as logistic regression. Used in the HASOC 2021 challenge \cite{hasoc2021mergeoverview}, it reached a medium performance level in English, the language for which it was easy to develop deep learning approaches relying on many external linguistic resources. Its performance, averaged on the two Hindi problems and the Marathi problem, ranks it in first place among the teams that proposed systems for at least two of these problems. These performances suggest that it is an interesting reference level to evaluate the benefits of using more complex approaches that are frequently used to address this type of task such as deep learning or taking into account complementary resources  \cite{DBLP:conf/fire/0001MMPDMP19,DBLP:conf/fire/0001MMC20,hasoc2021overview}. However, it is essential to note that the proposed system never ranked first in any specific task. It is therefore clearly not the best performing system for any of the five tasks.

\section*{Acknowledgments}
  The author wishes to thank the organizers of this shared task for putting together this valuable event and the reviewers for their very constructive comments. He is a Research Associate of the Fonds de la Recherche Scientifique - FNRS (F\'ed\'eration Wallonie Bruxelles de Belgique). Computational resources have been provided by the supercomputing facilities of the Universit\'e Catholique de Louvain (CISM/UCL) and the Consortium des Equipements de Calcul Intensif en F\'ed\'eration Wallonie Bruxelles (CECI).

\bibliography{eacl2017}
\bibliographystyle{eacl2017}

\end{document}